\documentclass{article}

\PassOptionsToPackage{numbers, compress}{natbib}
\usepackage{natbib}
\usepackage[preprint]{neurips_2025}

\usepackage{graphicx} 
\usepackage[most]{tcolorbox}
\usepackage{enumitem}
\usepackage{listings}
\usepackage{graphicx}
\usepackage{fancyvrb}
\usepackage{etoolbox}
\usepackage{graphicx}
\usepackage{fancyvrb}
\usepackage{dashrule}
\usepackage{authblk}
\usepackage{xcolor}
\usepackage{etoolbox}
\usepackage{inconsolata} 
\usepackage{listings}
\usepackage{booktabs}
\usepackage{algorithm}
\usepackage{algpseudocode}
\usepackage{amssymb}
\usepackage{hyperref}
\usepackage{inconsolata}   
\usepackage[T1]{fontenc}

\title{Omne-R1: Learning to Reason with Memory for Multi-hop Question Answering}
\author{Boyuan Liu, Feng Ji \thanks{Corresponding author: jifeng@tanka.ai}, Jiayan Nan, Han Zhao, Weiling Chen, Shihao Xu, Xing Zhou\\Tanka AI team, Tanka Inc.}

\begin{document}

\maketitle

\begin{figure}[htbp]
    \centering
    \includegraphics[width=0.9\textwidth]{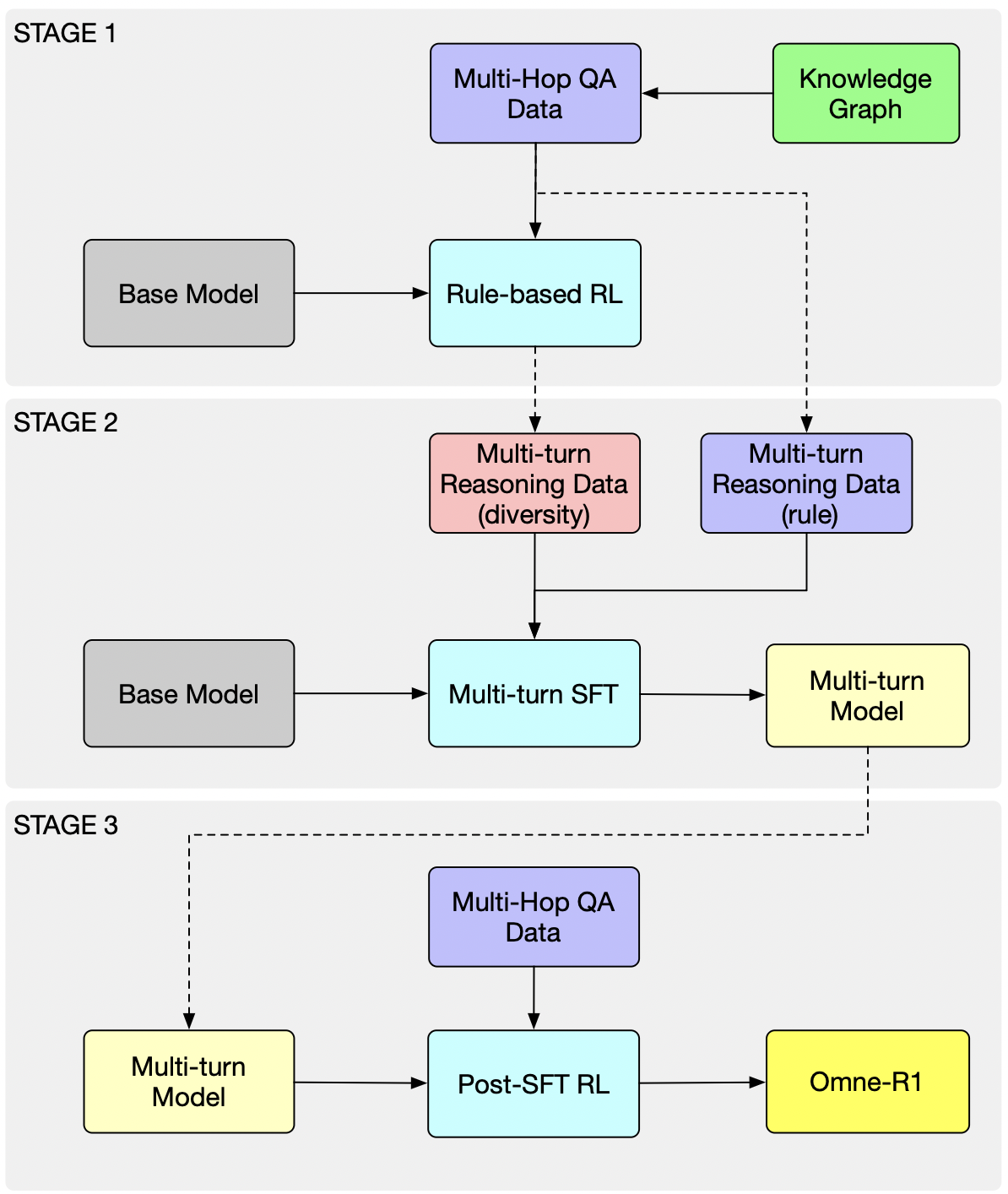} 
    \caption{Multi-Stage Training Workflow}
    \label{fig:workflow}
\end{figure}

\begin{abstract}
This paper introduces Omne-R1, a novel approach designed to enhance multi-hop question answering capabilities on schema-free knowledge graphs by integrating advanced reasoning models. Our method employs a multi-stage training workflow, including two reinforcement learning phases and one supervised fine-tuning phase. We address the challenge of limited suitable knowledge graphs and QA data by constructing domain-independent knowledge graphs and auto-generating QA pairs. Experimental results show significant improvements in answering multi-hop questions, with notable performance gains on more complex 3+ hop questions. Our proposed training framework demonstrates strong generalization abilities across diverse knowledge domains.
\end{abstract}

\section{Introduction}
Recently, research on large reasoning models \cite{deepseekai2025deepseekr1} \cite{kimiteam2025kimik15scalingreinforcement} \cite{xu2025largereasoningmodelssurvey} has advanced significantly, and subsequently, various reproduction efforts and reports \cite{openr1} \cite{tinyzero} \cite{zeng2025simplerlzooinvestigatingtamingzero} \cite{wen2025light} \cite{xie2025logicrlunleashingllmreasoning} \cite{hu2025openreasonerzeroopensourceapproach} have emerged over the past few months. Most focus on enhancing the reasoning capabilities in the field of mathematics or coding, since rule-based rewards can accurately measure results and enable considerable performance improvements via reinforcement learning. However, many natural language tasks, such as multi-hop question answering, also require strong reasoning abilities and need further exploration.

In this paper, we aim to solve multi-hop question answering on schema-free knowledge graphs by integrating the reasoning abilities of large reasoning models. Over the past decade, research in this field has predominantly followed two main approaches. The first approach focuses on developing semantic parsers that can translate natural language into structured query languages \cite{gan2021naturalsqlmakingsql} \cite{ma2025sqlr1trainingnaturallanguage} \cite{liu2025surveynl2sqllargelanguage}, such as Cypher \cite{ozsoy2024text2cypherbridgingnaturallanguage} or SPARQL \cite{liu2025nli4dbsystematicreviewnatural}, to query knowledge graphs. The second approach \cite{xue2022globalwalklearningglobalawarenode} \cite{Perozzi_2014} involves performing random walks on knowledge graphs based on entities and relationships identified in user queries. Our proposed approach in this paper belongs to the latter by simulating the traversal process from the question entity to the answer entity. Specifically, we harness the powerful reasoning capabilities of large language models to perform step-by-step multi-turn reasoning in a chain-of-thought format to complete the traversal and find the final answer, where each step sequentially performs a reasoning-query-acquire process.

To achieve the capability of answering multi-hop questions by traversing knowledge graphs, we propose a multi-stage training workflow, as shown in Figure~\ref{fig:workflow}. Initially, despite our extensive survey of various public knowledge graphs and question-answering datasets, the first problem that hindered our progress was the lack of suitable knowledge graphs and QA data tailored to our research goal. To address this, we first adopt an open-source project which is LightRAG\cite{guo2025lightragsimplefastretrievalaugmented} to construct multiple knowledge graphs. These graphs, cost-effective and automatically constructible, are schema-free and serve knowledge indexing purposes. Based on these graphs, we then design a method for utilizing a large reasoning model to automatically generate several tens of thousands of multi-hop QA pairs. This lays the foundation for the subsequent multi-stage model training process.

Our multi-stage model training workflow, shown in Figure~\ref{fig:workflow}, is composed of two reinforcement learning processes and one supervised fine-tuning process. In the first stage, we train the base model using a reinforcement learning algorithm with a rule-based reward. This stage enables the base model to understand and align with instructions and to answer some of the questions correctly. Only QA pairs are provided as training data in this stage. Subsequently, we employ the trained model to generate multi-turn reasoning QA data. Meanwhile, leveraging our prior knowledge of the original QA pair generation methodology, we also synthesize multi-turn reasoning QA data by constructing a trajectory template and then filling in the detailed reasoning content with the help of a large language model. Both datasets are utilized in the second stage of training. We adopt this hybrid dataset because relying solely on the latter data may lead to a bias toward a singular reasoning pattern, while the former data, with its diversity, benefits the generalization of reasoning capability. The hybrid data, along with the reasoning trajectories, is transformed into the multi-turn data format proposed by ShareGPT. After the second stage of supervised fine-tuning, we obtain a high-performance reasoning model capable of multi-hop question answering on knowledge graphs. To further enhance and explore the reasoning capability, we introduce the third stage: a post-SFT reinforcement learning process. As the fine-tuned reasoning model already has robust format-following capabilities, at this stage, we simplify the formatting reward function and redesign the rule-based reward method. At the end of the training workflow, we obtain the final model, named Omne-R1. We conduct experiments across all training stages and evaluate the resulting models and data. The analysis of the experimental results demonstrates the effectiveness of our proposed training workflow.

The remaining part of the paper is organized as follows:
Section~\ref{sec_data} details the methods of synthesizing multi-hop question-answer data based on knowledge graphs. Section~\ref{sec:tools} describes two essential tools designed to traverse knowledge graphs. Sections~\ref{sec_stage1}, \ref{sec_stage2}, and \ref{sec_stage3} present the three different stages of the proposed training workflow. Section~\ref{sec_exps} shows the experiments compared to two baselines. Finally, we present related works in Section~\ref{sec_rel} and the conclusion in Section~\ref{sec_conc}.

\section{Synthesizing Multi-hop QA Data from Knowledge Graphs}
\label{sec_data}

In order to train an LLM on the task of reasoning over a knowledge graph, a dataset of task descriptions with ground truth answers is essential for the training process. We generated approximately 70k multi-hop question-answer pairs using DeepSeek-R1 \cite{deepseekai2025deepseekr1}, all based on knowledge graphs constructed using LightRAG \cite{guo2025lightragsimplefastretrievalaugmented} across the 20 domains in the Ultradomain dataset \cite{qian2025memoragboostinglongcontext}. Compared with using only 4 domains as done in LightRAG, our use of 20 domains enhances the LLM's generalization capability across diverse knowledge graph types.

\subsection*{Multi-hop Question Generation}

To differentiate the difficulty levels of multi-hop questions, the generation process is divided into two categories: 1/2 hop questions and 3+ hop questions. To effectively train the LLM’s reasoning capability, the generated questions are required to meet the following criteria:

\textbf{Knowledge graph grounded}: The question must be answerable by querying the corresponding knowledge graph; in other words, there must exist a connected path within the graph that corresponds to the question.
\textbf{Non-trivial}: The question must require a non-trivial amount of KG querying and reasoning to obtain the answer. This implies that no node names, except for the starting node in the path, should be revealed, as this could leak a shortcut path to the answerer.
In contrast to related works such as Search-R1\cite{jin2025searchr1trainingllmsreason} and DeepResearcher\cite{zheng2025deepresearcherscalingdeepresearch}—where answer traceability through the search tool is not always guaranteed—our requirement for grounding the question in the knowledge graph helps reduce ambiguity and improves the training signal.

For both 1/2 hop and 3+ hop questions, the generation procedure is as follows:

\textbf{Step 1}: For a given knowledge graph, we sample a number of seed nodes based on the following criteria:
\begin{itemize}
\item There exists a subgraph originating from the seed node with a depth of 3, discovered via breadth-first search.
\item The total number of nodes in this subgraph should be within a reasonable range, typically from 20 to 400.
\end{itemize}

\textbf{Step 2}: For each seed node, we extract a subgraph of depth 3 and present it to an LLM to generate: (i) a selected path within the subgraph considered reasonable, (ii) a multi-hop question, and (iii) a node name as the golden answer. The LLM used in this step is DeepSeek-R1. For 1/2 hop question generation, we incorporate few-shot examples from HotpotQA \cite{yang2018hotpotqadatasetdiverseexplainable} in the prompt to enhance question quality and diversity. Previous generated paths, questions, and answers may be included in the prompt to improve the result if the first attempt fails validation (explained below). The complete generation prompt is available in the Appendix \ref{sec:question_generation_prompts}, with optional reused content shown in gray.

\textbf{Step 3}: We evaluate both the selected path and the generated question to ensure quality. For path validation, we consider a path valid if it contains the answer node and is a connected path in the knowledge graph. For question validation, we provide both the question and the path to the LLM, which is prompted to assess validity based on defined criteria and explain its judgment. DeepSeek-R1 is also used for this step. The evaluation prompt is detailed in the Appendix \ref{sec:question_generation_prompts}.

Here is an example for the multi-hop question generation result.

\begin{tcolorbox}[colback=gray!5!white, colframe=gray!80!black, title=Multi-hop question example, fonttitle=\bfseries]
\textbf{1/2 hop:}

path: Colony Collapse Disorder (CCD) $\rightarrow$ Nosema $\rightarrow$ Fumagilin-B

question: What biochemical intervention is used to manage the pathogen that contributes to Colony Collapse Disorder through pesticide interactions?

golden answer: Fumagilin-B

\vspace{1em}

\textbf{3+ hop:}

path: Madame de Merret $\rightarrow$ La Grande Breteche $\rightarrow$ Vendome $\rightarrow$ Monsieur Regnault $\rightarrow$ Comtesse de Merret

question: Madame de Merret owned a property that was central to mysterious posthumous wishes regarding its preservation. This estate held geographical significance in a town connected to inheritance investigations. A legal professional relocated to this town through family connections and later handled legal enforcement of will stipulations for another aristocratic figure. Who inherited the property through these estate administration procedures?

golden answer: Comtesse de Merret

\end{tcolorbox}

It should be noted that the correctness of each question-answer pair depends on the specific knowledge graph. In other words, an answer is considered correct only within the context of that particular graph.

In total, we have generated around 20k 1/2 hop questions and 50k 3+ hop questions. We have de-duplicated the dataset by ensuring uniqueness over question-answer pairs. The statistics of multi-hop questions for each knowledge domain are summarized in Table~\ref{tab:category_hop_distribution}. Due to the varying connectivity properties of the knowledge graphs, the number of questions per domain varies significantly.

\begin{table}[ht]
\centering
\begin{tabular}{lrrr}
\hline
\textbf{Graph Type} & \textbf{1/2 hop} & \textbf{3+ hop} & \textbf{Total} \\
\hline
agriculture  & 286  & 2993 & 3279 \\
cs           & 888  & 1982 & 2870 \\
legal        & 357  & 929  & 1286 \\
mix          & 699  & 1753 & 2452 \\
biology      & 1552 & 4760 & 6312 \\
history      & 1240 & 5333 & 6573 \\
health       & 1287 & 4029 & 5316 \\
art          & 912  & 3731 & 4643 \\
cooking      & 806  & 1988 & 2794 \\
physics      & 1212 & 1679 & 2891 \\
fiction      & 1127 & 2759 & 3886 \\
literature   & 1416 & 3389 & 4805 \\
music        & 887  & 1350 & 2237 \\
psychology   & 1228 & 3097 & 4325 \\
biography    & 912  & 1856 & 2768 \\
fin          & 710  & 1490 & 2200 \\
mathematics  & 1461 & 3101 & 4562 \\
philosophy   & 1499 & 2345 & 3844 \\
politics     & 1034 & 2220 & 3254 \\
technology   & 2021 & 1309 & 3330 \\
\hline
Total        & 21534 & 52093 & 73627 \\
\hline
\end{tabular}
\vspace{0.5em}
\caption{Distribution of 1/2 hop and 3+ hop questions by domain}
\label{tab:category_hop_distribution}
\end{table}

\section{Multi-Stage Training Workflow}

    The proposed multi-stage training workflow is comprehensively described in this section. We start with the detailed descriptions of multi-turn reasoning loop in section \ref{sec:loop}, and two traversal tools on knowledge graphs in section \ref{sec:tools}. Then the whole training workflow is sequentially presented in section \ref{sec_stage1}, \ref{sec_stage2} and \ref{sec_stage3}.

\subsection{Multi-turn Reasoning Loop on Knowledge Graphs}
\label{sec:loop}

Our LLM is designed to interact with the knowledge graph in an iterative manner until the final answer is reached. Specifically, the output of the LLM falls into one of the following two categories:

\textbf{Category 1}: Interact with the knowledge graph by triggering a tool call. The LLM should first output a reasoning process enclosed within \texttt{<think>} and \texttt{</think>} tags, followed by the tool call specification enclosed within \texttt{<tool\_call>} and \texttt{</tool\_call>} tags.

\textbf{Category 2}: Provide the final answer. The LLM should first output a reasoning process enclosed within \texttt{<think>} and \texttt{</think>} tags, and then provide the answer enclosed within \texttt{<answer>} and \texttt{</answer>} tags.

If the LLM’s output falls into Category 1, the tool call result will be wrapped within \texttt{<tool\_response>} and \texttt{</tool\_response>} tags and appended to the current prompt as a user message for the next generation round. If the output falls into Category 2, the final answer will be extracted from the content between the \texttt{<answer>} and \texttt{</answer>} tags, and the reasoning loop will terminate.

In our setting, the maximum number of reasoning rounds (i.e., tool calls) is capped at 7. If this limit is exceeded, the generation loop is forcibly terminated. Each round of LLM generation is restricted to a maximum of 3000 tokens, including the reasoning content.

\subsection{Tool Descriptions}
\label{sec:tools}

To enable the LLM to effectively navigate and utilize the extensive knowledge graph, a set of essential tools is provided. These tools are designed to support search and traversal operations, equipping the LLM with capabilities comparable to a human exploring a graph through querying and reasoning. We provide two main tools for interacting with the knowledge graph:

\textbf{entity matcher}: Given a search query and the graph type, this tool returns a list of matching node names along with their descriptions. The returned nodes are sorted by similarity to the query in descending order. This tool does not include edge or neighbor information and is primarily used to retrieve exact node names through semantic search.

\textbf{node info}: Given a specific node name and graph type, this tool returns detailed information about the node, including the names of all neighboring nodes and the descriptions of their connecting edges. This tool is essential for understanding node relationships and enabling traversal from one node to its neighbors.

The graph type is specified in the system prompt, and the LLM is expected to fill in the correct graph type in tool calls in order to produce valid answers. Refer to Appendix section \ref{sec:systemp_prompt} for detailed system prompt. For a complete tool usage examples and interaction traces, refer to the Appendix section \ref{sec:sft-data-preparation}.

\subsubsection{Implementation Details}

During the training of LLM reasoning over knowledge graphs—particularly in the reinforcement learning stage where multiple rollouts are required—efficient and scalable handling of tool calls is critical for maintaining training throughput.

We implement multiple web services using FastAPI \cite{Ramirez_FastAPI_2025} to expose convenient endpoints for tool function calls. For the \textbf{entity matcher} tool, considering the balance between semantic matching quality and response time, we use the embedding model \texttt{bge-large-en-v1.5} \cite{bge_embedding} to embed both the query and entity names. Embeddings for all entities across the 20 knowledge graphs are precomputed and stored in NanoVectorDB \cite{Ye_2024_nanoVectordb}, a vector database that supports fast approximate nearest-neighbor retrieval based on cosine similarity.

For the \textbf{node info} tool, which returns neighbor node information and edge details, we use NetworkX \cite{Hagberg_NetworkX_2008} to manage and query the knowledge graphs. The raw outputs from both tools are post-processed and rendered with structured templates to improve readability for both the LLM and human.

\subsection{Stage 1: Rule-based RL}
\label{sec_stage1}

In this study, we begin by training a model using rule-based reinforcement learning (RL) to generate reasoning steps as a foundation for subsequent supervised fine-tuning (SFT).

\subsubsection{Methodology}

The training process is based on the Proximal Policy Optimization (PPO) algorithm, which optimizes the policy $\pi_\theta$ by maximizing the expected reward while constraining policy updates to remain close to the previous policy $\pi_{\theta_{\text{old}}}$. The PPO objective function is defined as:

$$
L^{\text{PPO}}(\theta) = \mathbb{E}_t \left[ \min \left( r_t(\theta) \hat{A}_t, \text{clip}(r_t(\theta), 1 - \epsilon, 1 + \epsilon) \hat{A}_t \right) \right]
$$

where $r_t(\theta) = \frac{\pi_\theta(a_t|s_t)}{\pi_{\theta_{\text{old}}}(a_t|s_t)}$ is the ratio of the probabilities assigned by the new and old policies, and $\hat{A}_t$ is the estimated advantage at time step $t$. The clipping term helps to avoid overly aggressive updates, controlled by the hyperparameter $\epsilon$.

The advantage $\hat{A}_t$ is computed using Generalized Advantage Estimation (GAE):

$$
\hat{A}_t = \delta_t + (\gamma \lambda) \delta_{t+1} + \dots + (\gamma \lambda)^{T - t + 1} \delta_{T-1}
$$

with $\delta_t = r_t + \gamma V(s_{t+1}) - V(s_t)$, where $V(s_t)$ is the value function estimating the expected return from state $s_t$, $\gamma$ is the discount factor, and $\lambda$ is the GAE parameter controlling the bias-variance trade-off.

We adopt an actor-critic architecture composed of two neural networks: the actor $\pi_\theta$, which outputs a probability distribution over actions given a state, and the critic $V_\phi$, which estimates the value function. Both networks share an encoder that processes the input prompts and the model's previously generated outputs. During training, the actor is updated to maximize the PPO objective, while the critic minimizes the mean squared error between its predictions and the observed returns:

$$
L^{\text{value}}(\phi) = \mathbb{E}_t \left[ \left( V_\phi(s_t) - R_t \right)^2 \right]
$$

where $R_t$ is the cumulative reward obtained from time step $t$ onward.

To ensure training stability and efficiency, we use a replay buffer and perform mini-batch updates. The model generates multiple trajectories, each consisting of a sequence of states, actions, rewards, and value estimates. These trajectories are used to compute both the PPO loss and the value loss, which are then optimized via stochastic gradient descent.

\subsubsection{Rule-based Reward}

Since the answers to multi-hop questions are entities that exist in the knowledge graphs, exact match evaluation is a suitable approach for judging the correctness of final answers. However, our initial experiments revealed that the model exhibited a strong tendency to directly invoke traversal tools without conducting proper reasoning first. A similar issue is discussed in \cite{jin2025empiricalstudyreinforcementlearning}. Given the importance of the reasoning process—not only for improving model capability but also for enhancing interpretability—we introduce a hierarchical reward structure that incorporates both formatting discipline and outcome correctness.

The total reward $R_{\text{total}}$ is defined as:

$$
R_{\text{total}} = R_{\text{format}} \times (0.1 + 0.9 \times R_{\text{result}})
$$

Here, $R_{\text{format}} \in [0, 1]$ measures the compliance of the model’s output with the predefined structured format (i.e., correct use of \texttt{<think>}, \texttt{<tool\_call>}, and related tags). Higher values correspond to better format adherence. $R_{\text{result}} \in \{0, 1\}$ reflects whether the final answer matches the ground truth exactly—1 for correct, 0 for incorrect.

This reward scheme ensures that while correctness of the final answer is prioritized, the structural integrity of the reasoning steps is also encouraged, helping the model develop clearer and more interpretable reasoning behavior.

\subsection{Stage 2: Multi-turn SFT}
\label{sec_stage2}

Leveraging the data generated from the reinforcement learning phase as described previously, we constructed a multi-turn SFT dataset. Each interaction within this dataset is meticulously designed to encompass the complete Chain-of-Thought (CoT) process, the corresponding tool call, and the response from the Knowledge Graph (KG) query interface. This structure ensures that the model is trained on complete and correct reasoning trajectories. The multi-turn dialogues were formatted using a ShareGPT-style structure, adapted to incorporate tool calls effectively. Detailed examples and the specific construction methodology are provided in Appendix \ref{sec:sft-data-preparation}.

\subsubsection{Training Methodology}
The SFT process was conducted using the LLaMA Factory framework\cite{zheng2024llamafactoryunifiedefficientfinetuning}. We selected models from the recently proposed Qwen3 series as the base for our fine-tuning efforts. The Qwen3 family of models, as detailed in their technical report \cite{yang2025qwen3}, demonstrates strong performance in complex reasoning and tool utilization tasks. A key feature of Qwen3 is its integrated operation with both a \textbf{thinking mode} and a \textbf{non-thinking mode}. This is achieved by training the model on a mixture of data that includes explicit reasoning steps and data that does not. During inference, if thinking mode is active, the model first generates its reasoning process encapsulated within \texttt{<think>} and \texttt{</think>} tags before producing the final output. Conversely, in non-thinking mode, an empty CoT (e.g., \texttt{<think>\textbackslash n\textbackslash n</think>}
) is prepended to the input to encourage faster, more direct responses. This paradigm allows for a flexible balance between deliberative, slow thinking and rapid, reactive responses, adapting to the complexity of the input query.

To maintain consistency with the Qwen3 architecture and optimize for its capabilities, we aligned our SFT data preparation with its specific prompting and tool use templates. Tool calls are within \texttt{<tool\_call>} and \texttt{</tool\_call>} tags. 

The subsequent responses from our KG query interface, representing the outcome of the tool execution, are similarly enclosed in \texttt{<tool\_response>} and \texttt{</tool\_response>} tags. This alignment ensures that the model learns to process and generate tool interactions in the format it expects. 


\subsubsection{Hyperparameter Configuration}

The fine-tuning was performed on the Qwen3-14B model with both full-parameter setting and Low-Rank Adaptation (LoRA) \cite{hu2022lora} setting. For LoRA fine-tuning, we employed a LoRA rank of 8 applied to all linear layers. The maximum sequence length was set to 6144 tokens. 
The model was trained for 3 epochs using a cosine learning rate scheduler, with a learning rate of \(1.0 \times 10^{-4}\) and a warmup ratio of 0.1. For Full SFT, the learning rate and warmup ratio was set to \(1.0 \times 10^{-5}\) and 0.01. 
An effective batch size of 4 was used, achieved through a per-device batch size of 1 and 4 gradient accumulation steps. Training was conducted with bfloat16 precision, leveraging DeepSpeed ZeRO Stage 0 for efficient distributed training.

\subsubsection{Weighted Loss Masking for thinking and tool call tokens}

During our SFT phase, the model is trained on data where each assistant turn can contain both a CoT block (enclosed in \texttt{<think>} and \texttt{</think>} tags) and a subsequent tool call. A standard cross-entropy loss would treat all tokens within this turn equally. This can inadvertently lead the model to prioritize the generation of extensive CoT narratives, potentially at the expense of accurately formulating tool calls. For instance, the model might produce overly verbose reasoning steps or, more critically, make errors in the parameters of a tool call, such as using an incorrect node name.

To mitigate this, we investigated the use of a weighted loss mask. The core idea is to selectively down-weight the contribution of tokens within the CoT blocks to the overall loss. By doing so, we aim to reduce the penalty associated with deviations in the CoT content, thereby relatively emphasizing the importance of other parts of the generation, particularly the correctness of the tool calls.

We implemented a custom loss computation within our training loop. The procedure involves identifying the token spans corresponding to the CoT (i.e., tokens between \texttt{<think>} and \texttt{</think>} tags) and applying a significantly smaller weight to these tokens compared to others. The procedure for this modified loss calculation is outlined in Algorithm~\ref{alg:weighted_loss}. 

\begin{algorithm}[h!]

\renewcommand{\algorithmicrequire}{\textbf{Input:}}
\renewcommand{\algorithmicensure}{\textbf{Output:}}

\caption{Weighted Loss Computation for SFT with Thinking Mode}
\label{alg:weighted_loss}
\begin{algorithmic}[1]
\Require Model $M$, Input data $I$ (containing labels $L$)
\Ensure Computed weighted loss value $\mathcal{L}_{final}$

\State $(\text{logits}, \dots) \gets M(I)$
\State $L \gets I[\text{"labels"}]$
\State $\text{weight\_mask} \gets \text{Tensor filled with 1s, shape of } L$

\State $\text{think\_start\_id} \gets \text{TokenID}(\texttt{<think>})$
\State $\text{think\_end\_id} \gets \text{TokenID}(\texttt{</think>})$

\For{$i = 0 \to \text{batch\_size} - 1$}
    \State $\text{starts} \gets \text{FindIndices}(L[i], \text{think\_start\_id})$
    \For{\textbf{each} $s \in \text{starts}$}
        \State $\text{ends\_relative} \gets \text{FindIndices}(L[i, s:], \text{think\_end\_id})$
        \If{$\text{ends\_relative is not empty}$}
            \State $e \gets s + \text{ends\_relative}[0]$
            \State $\text{weight\_mask}[i, s:e+1] \gets 0.1$ \Comment{Apply small weight}
        \EndIf
    \EndFor
\EndFor

\State $\mathcal{L}_{CE} \gets \text{CrossEntropyLossFunction}(\text{reduction='none'})$

\State $\text{shift\_logits} \gets \text{logits}[:, :-1, :]$
\State $\text{shift\_labels} \gets L[:, 1:]$
\State $\text{shift\_weights} \gets \text{weight\_mask}[:, 1:]$

\State $\text{loss\_per\_token} \gets \mathcal{L}_{CE}(\text{shift\_logits}, \text{shift\_labels})$
\State $\text{weighted\_loss} \gets \text{loss\_per\_token} \times \text{shift\_weights}$

\State $\text{active\_mask} \gets (\text{shift\_labels} \neq \text{IGNORE\_INDEX})$
\State $\text{masked\_weighted\_loss} \gets \text{weighted\_loss} \times \text{active\_mask}$

\State $\text{effective\_sum} \gets \sum (\text{shift\_weights} \times \text{active\_mask})$
\If{$\text{effective\_sum} > 1e-8$}
    \State $\mathcal{L}_{final} \gets (\sum \text{masked\_weighted\_loss}) / \text{effective\_sum}$
\Else
    \State $\mathcal{L}_{final} \gets 0$
\EndIf

\State \Return $\mathcal{L}_{final}$
\end{algorithmic}
\end{algorithm}

By assigning a substantially smaller weight (e.g., 0.1, as indicated in Algorithm~\ref{alg:weighted_loss}) to the tokens within the \texttt{<think>}...\texttt{</think>} blocks, their contribution to the overall loss value is significantly diminished. Consequently, during the backpropagation phase, the gradients originating from these CoT tokens are proportionally scaled down. This scaling occurs because the gradient of the loss with respect to any model parameter is a sum of terms, where each term related to a specific token's prediction error is multiplied by that token's loss weight. A smaller weight directly reduces the magnitude of these gradient components. This does not mean these tokens are completely ignored (unless the weight is zero), but rather that the model's parameters are updated less aggressively based on errors within the CoT content. The effect is to de-emphasize rote memorization or overly precise replication of the thinking steps, while relatively increasing the learning signal from other, unweighted (or higher-weighted) parts of the output, such as the tool call formulation and the final answer. Finally, the sum of weights for all non-padded tokens is calculated to ensure the overall loss magnitude remains comparable across different examples or batches, preventing training instability that might arise if the denominator were simply the count of active tokens rather than their weighted sum.

\subsection{Stage 3: Post-SFT RL}
\label{sec_stage3}

We apply reinforcement learning to further refine the model trained via SFT in Stage 2. Given that the SFT-trained model has already learned to imitate the tool-calling trace and reasoning structure to a reasonable extent, reinforcement learning provides an opportunity to optimize the model's behavior by actively exploring improved reasoning trajectories that maximize a custom reward function.

Specifically, we conducted PPO training on two models from Stage 2:

1. The Qwen3-14B model trained via SFT with loss weighted mask factor equal to 0.001.

2. The Qwen3-14B model trained via full-parameter SFT in no-think mode.

To further boost the model’s multi-hop reasoning ability and discourage undesirable behaviors, we designed a modified reward function defined as:

\begin{align*}
    R = (r_{format} + r_{answer}) / 2 - n_{repetition} \cdot 0.1
\end{align*}

where the format score $r_{format}$ and answer score $r_{answer}$ is defined as follows,
\begin{align*}
    r_{\text{format}} &=
    \begin{cases}
        1 & \text{if the format is correct} \\
        0 & \text{otherwise}
    \end{cases}
\end{align*}

\begin{align*}
    r_{\text{answer}} &=
    \begin{cases}
        1 & \text{if the answer is correct} \\
        0 & \text{otherwise}
    \end{cases}
\end{align*}

In addition to these standard evaluation signals, we introduce a repetition penalty term to discourage the model from issuing redundant tool calls. The term $n_{\text{repetition}}$ counts the number of duplicate tool calls—those that invoke the same function name with identical input parameters. Such repetitions contribute no new information, as their results are already available in the existing dialogue context. Repeated tool calls not only increase unnecessary context length but also waste tool-calling rounds, which are strictly limited in our setup.

After applying reinforcement learning with the above reward, we observed accuracy improvements on both models originating from Stage 2, demonstrating the effectiveness of this post-SFT RL refinement in further enhancing multi-hop reasoning performance.

\section{Experiments}
\label{sec_exps}

\subsection{Evaluation}
To assess the capabilities of our trained language models on multi-hop question answering with knowledge graph, we conducted evaluations on subsets of the dataset describe in Appendix \ref{sec:sft-data-preparation}:

\begin{itemize}
    \item \textbf{1/2 hop dataset}: Consists of 1106 questions that require 1 or 2 hops of reasoning over the knowledge graph.
    \item \textbf{3+ hop dataset}: Contains 1453 questions involving deeper reasoning chains requiring 3 or more hops.
\end{itemize}

\subsubsection{Evaluation Setup}

We use a custom inference pipeline built on top of the vLLM engine to generate answers in batches of 32 questions. For each question, the model follows an iterative reasoning process guided by structured prompts.

In particular, we employ a \textbf{think mode} prompting strategy. Under this setup, the model is instructed to explicitly think between reasoning steps using special \texttt{\textless think\textgreater} and \texttt{\textless/think\textgreater} tags. At each round, the model produces a thought segment and can optionally trigger a tool call to query the knowledge graph or provide a final answer as described in section \ref{sec:loop}.


The reasoning continues for a maximum of 7 rounds or until the model emits a final answer enclosed in \texttt{\textless answer\textgreater} and \texttt{\textless/answer\textgreater} tags.

\subsubsection{Accuracy Computation}

The final predicted answer for each question is extracted from the last \texttt{\textless answer\textgreater} block in the generated dialog. A prediction is considered correct if it matches the gold answer exactly (after whitespace stripping).

Overall accuracy is computed as:
\[
\text{Accuracy} = \frac{\text{\# correct predictions}}{\text{\# total questions}}
\]

\subsection{Analysis}

\subsubsection{Experiments on baseline models/agent}

We conducted evaluations on the base models to establish baseline accuracy. To assess the inherent capabilities of the base model both with and without access to knowledge graph (KG) tools, we included a setting where the model directly answers the questions without any external assistance, thereby testing its innate knowledge. The base model here is Qwen3-14B. The detailed results are in Table \ref{tab:baseline_result}.

We also evaluated a retrieval-augmented generation (RAG) framework—LightRAG \cite{guo2025lightragsimplefastretrievalaugmented}—on our multi-hop questions. For consistency, we replaced the original base model in LightRAG with Qwen3-14B and substituted the embedding model with \texttt{bge-large-en-v1.5}, to align it with the configuration used in our KG tool-based reasoning experiments. Since Qwen3-14B has both think and no think mode, both two modes are tested.

Since neither the direct answering nor the LightRAG setting enforces a fixed answer format, we employed LLM-based evaluation using DeepSeek-R1. In this evaluation, the model is provided with the question, the golden answer, and the predicted answer, and is asked to judge whether the answer is correct, providing reasoning for its decision. This evaluation relies solely on the LLM's internal judgment, without additional prompting or constraints.

In contrast, for evaluating LLM reasoning with KG tool interactions, we use exact match (EM) as the evaluation metric as mentioned above, given that the answer format is explicitly defined and enforced in these settings. It is important to note that results from EM and LLM-based evaluation are not directly comparable. EM is stricter, and deviations in formatting—rather than semantic correctness—can lead to lower scores. 

\begin{table}[htbp]
\centering
\begin{tabular}{@{}cccc@{}}
\toprule
\textbf{Model/Agent} & \textbf{1/2 Hop (\%)} & \textbf{3+ Hop (\%)} & \textbf{Evaluation method} \\
\midrule
Direct answer(w think) & 26.0 & 26.3 & R1 \\
Direct answer(w/o think) & 23.8 & 26.2 & R1 \\
LightRAG(w think) & 62.5 & 49.5 & R1 \\
LightRAG(w/o think) & 59.2 & 46.8 & R1 \\
Base Model(w think) reasoning with KG tools & 34.9 & 15.6 & EM \\
\bottomrule
\end{tabular}
\vspace{0.5em}
\caption{Baseline evaluation results. The base model is Qwen3-14B.}
\label{tab:baseline_result}
\end{table}

\subsubsection{Experiments of Stage 1}

The experiments of stage 1 are conducted by utilizing a reinforcement learning framework along with a rule-based reward, which details can be found in Section \ref{sec_stage1}. Since Qwen3 series models had not been released when we trained our model, we opted for the Qwen2.5-14B series as the initial model. It should be noted that we first conducted separate experiments on 1/2-hop data and 3+ hop data. Although we also tried with the mix data, the results showed difference.

\begin{table}[htbp]
\centering
\begin{tabular}{@{}cccc@{}}
\toprule
\textbf{Method} & \textbf{Train Data} & \textbf{1/2 Hop (\%)} & \textbf{3+ Hop (\%)} \\
\midrule
Rule-based RL & 1/2 Hop & 71.2 & -  \\
Rule-based RL & 3+ Hop & - & 28.9 \\
Rule-based RL & mix & 57.5 & 32.1 \\
\bottomrule
\end{tabular}

\vspace{0.5em} 
\caption{Accuracy of multi-hop QA on Qwen2.5-14B-Instruct in Stage 1}
\label{tab:post_rl_stage1}
\end{table}

All experimental results in this stage are listed in Table \ref{tab:post_rl_stage1}. Although achieving the accuracy of 71.2\% on 1/2 hop data, the accuracy drops significantly to 28.9\% on 3+ hop data. We also attempted to enhance the model's reasoning ability through curriculum learning, but it was not successful. It is evident that solely training with reinforcement learning from a base model proves insufficient to enable the model's capability to traverse on knowledge graphs.

\subsubsection{Experiments of Stage 2}

Table~\ref{tab2} presents the performance of various SFT strategies on the Qwen3-14B model. Notably, both LoRA and Full SFT models show significant improvements over the original base model, confirming the effectiveness of supervised fine-tuning in enhancing reasoning performance.

Overall, models fine-tuned with full-parameter supervised fine-tuning (Full SFT) consistently outperform LoRA-based models, especially on the more complex 3+ hop questions, indicating that full fine-tuning offers stronger reasoning capabilities for multi-hop inference.

Interestingly, enabling the \texttt{think} mode—designed to promote deeper reasoning—tends to degrade performance across both LoRA and Full SFT settings. This suggests a possible incompatibility between manually prompted reasoning modes and the reasoning strategies learned during SFT.

To further investigate this observation, we designed a weight mask loss strategy to understand the contribution of reasoning and tool call part in the response of LLM in the next section.

\begin{table}[h]
\centering
\begin{tabular}{@{}cccc@{}}
\toprule
\textbf{Method} & \textbf{Deepthink} & \textbf{1/2 Hop (\%)} & \textbf{3+ Hop (\%)} \\ 
\midrule
No SFT       & Yes                & 34.9                 & 15.6                 \\
Lora         & No                 & 92.4                  & 60.0                 \\
Lora         & Yes                & 83.8                 & 47.2                 \\
Full         & No                 & 94.1                 & 68.4                 \\
Full         & Yes                & 89.2                 & 57.1   \\             
\bottomrule
\end{tabular}

\vspace{0.5em}
\caption{SFT Performance on Qwen3-14B}
\label{tab2}
\end{table}

\textbf{Impact of Weighted Loss Mask on Multi-hop Accuracy}

\begin{figure}[htbp]
    \centering
    \includegraphics[width=0.8\textwidth]{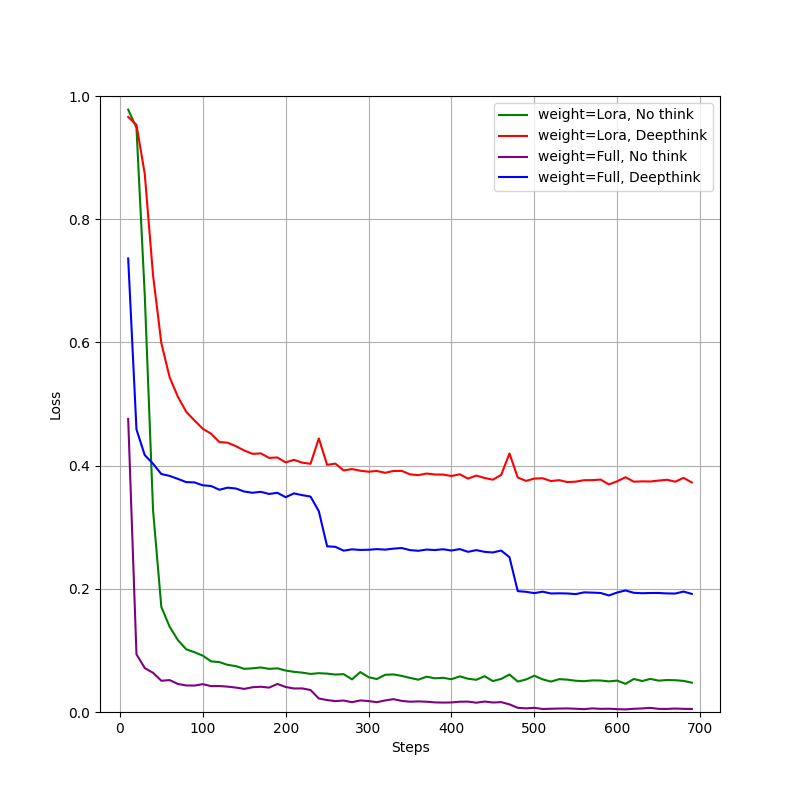}
    \caption{Training loss curves under different sft settings.}
    \label{fig:sft_loss_comparison}
\end{figure}

\begin{figure}[htbp]
    \centering
    \includegraphics[width=0.8\textwidth]{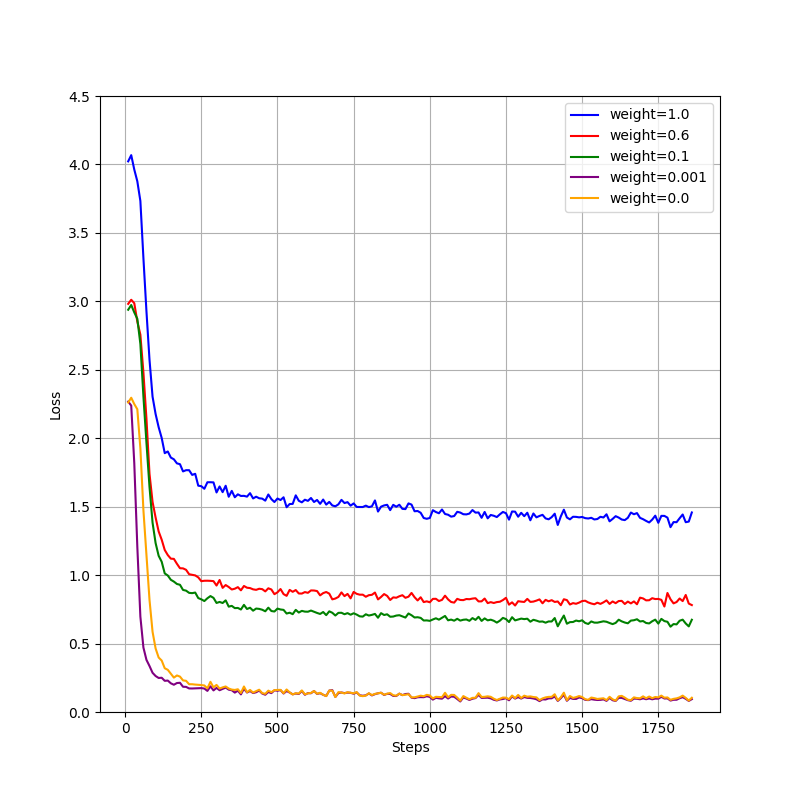}
    \caption{Training loss curves under different loss mask weight settings.}
    \label{fig:loss_comparison}
\end{figure}

To evaluate the influence of the thinking process on downstream task performance, we configured a series of experiments with varying loss mask weights and evaluated the model's accuracy on 1/2 hop and 3+ hop questions. The results are presented in Table \ref{tab:hop_accuracy}.

\begin{table}[htbp]
\centering
\begin{tabular}{@{}ccc@{}}
\toprule
\textbf{Weighted Loss Mask} & \textbf{1/2 Hop (\%)} & \textbf{3+ Hop (\%)} \\
\midrule
1.00  & 87.9 & 52.8 \\
0.80  & 88.0 & 53.8 \\
0.20  & 88.4 & 54.2 \\
0.10  & 87.9 & 53.4 \\
0.05  & 88.6 & 54.4 \\
0.01  & 88.6 & 55.6 \\
\textbf{0.001} & \textbf{89.2} & \textbf{60.5} \\
0.00  & 83.0 & 43.8 \\
\bottomrule
\end{tabular}

\vspace{0.5em} 
\caption{Accuracy on 1/2 hop and 3+ hop questions with different Loss Weight Mask settings.}
\label{tab:hop_accuracy}
\end{table}

As illustrated in Figure \ref{fig:sft_loss_comparison}, we can observe a clear trend in the training process. As we decrease the loss mask weight, the overall training loss also systematically decreases. This is the expected behavior because a smaller weight diminishes the loss contribution from the `think` part tokens and weakens their impact on gradient updates. When the final loss is computed as a weighted average, a lower weight on the think tokens naturally leads to a smaller overall training loss, confirming the effectiveness of our weighting mechanism.

A Weighted Loss Mask of 1.0 represents our baseline, equivalent SFT approach where the \texttt{<think>} and \texttt{<tool\_call>} tokens contribute equally to the loss computation. As shown in Table~\ref{tab:hop_accuracy}, the weight setting significantly impacts the model's accuracy. We observe a consistent improvement in performance as the weight is reduced from 1.0 down to 0.001. This trend suggests that by de-emphasizing the loss from the \texttt{<think>} portion, we allow the model to learn more generalizable reasoning patterns rather than overfitting to the specific verbiage of the thought process in our SFT dataset. The model achieves its peak performance at a weight of 0.001, with an accuracy of 89.2\% on 1/2 hop and a remarkable 60.5\% on 3+ hop questions.

However, as the weight decreases further from 0.001 to 0.0, a notable decline in accuracy occurs. Upon closer inspection of the model's behavior with a weight of 0.0, we found that its reasoning process reverted to a style closely resembling the native Qwen3-14B model. The model's outputs often began with verbose, generic phrases such as "Okay, let's try to figure this out step by step" or expressions of uncertainty like "Wait, but I'm not sure...". While this indicates a regression to its pretrained behavior, such a lengthy and sometimes hesitant thinking process can confuse the model and is not always optimal for the structured, multi-hop KG querying task.

In conclusion, our experiments identify a loss weight of 0.001 as the optimal setting for our task. This value strikes a critical balance: it is low enough to weaken the influence of the \texttt{<think>} tokens, preventing the model from merely mimicking patterns in the training data, yet it is non-zero, thus preserving the essential KG reasoning capabilities instilled during fine-tuning. This prevents the model from completely degrading to the original Qwen3's reasoning style, thereby making it significantly more effective and suitable for our specific multi-hop QA task.

\subsubsection{Experiments of Stage 3}

We continued to perform RL training to the models after Stage 2, aiming to achieve a higher accuracy on the multi-hop question solving with knowledge graph. As mentioned above, the PPO training is conducted on the two models from Stage 2: Qwen3-14B SFT(Lora) with loss weighted mask=0.001 and Qwen3-14B SFT(full) in no think mode.

\begin{figure}[htbp]
    \centering
    \includegraphics[width=0.8\textwidth]{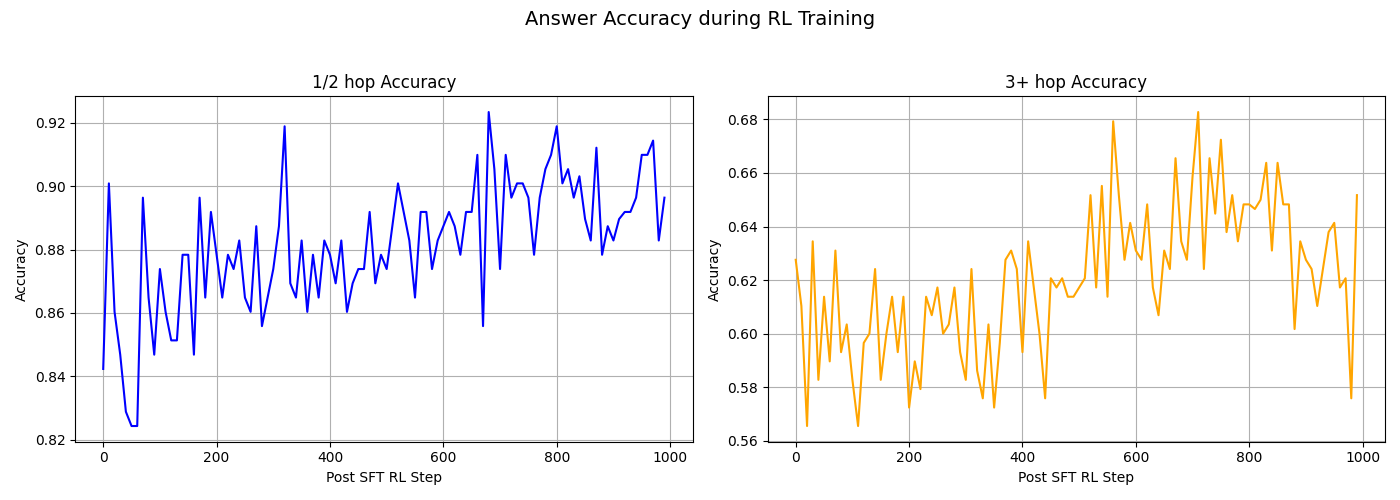}
    \caption{Accuracy curve during RL training on model after SFT(Lora) with loss weight mask=0.001, on a subset of size 512}
    \label{fig:post_sft_rl_accuracy_lwm1e-3_side_by_side}
\end{figure}

During the post training of RL, the trend of accuracy increase has been spotted for both models. For evaluation, we adopt the same inference backend as SFT of which the batch size of vllm inference is 32. The detailed accuracy after Post-SFT RL is shown in Table \ref{tab:post_sft_rl_hop_accuracy}. The plot of answer accuracy curve during RL training on the model after SFT(Lora) with loss weighted mask=0.001 is plotted in Figure\ref{fig:post_sft_rl_accuracy_lwm1e-3_side_by_side}, where the dataset for evaluation is a random size 512 subset of the test dataset, which may differ from the full test dataset evaluation result. 

\begin{table}[htbp]
\centering
\begin{tabular}{@{}lcc@{}}
\toprule
\textbf{Model} & \textbf{1/2 Hop (\%)} & \textbf{3+ Hop (\%)} \\
\midrule
SFT (LoRA) & 89.2 & 60.5 \\
+ RL (LoRA) & 92.7 & 63.5 \\
\midrule
SFT (Full) & 94.1 & 68.4 \\
+ RL (Full) & 94.0 & 69.3 \\
\bottomrule
\end{tabular}
\vspace{0.5em}
\caption{Accuracy of multi-hop QA in Stage 3. LoRA model uses a loss-weighted mask of 0.001. RL denotes PPO-based reinforcement learning applied post-SFT.}
\label{tab:post_sft_rl_hop_accuracy}
\end{table}

Moreover, a difference in terms of tool call trajectory pattern has been spotted before and after Stage 3 RL training. Before the Stage 3 RL training, trajectories of multiple \texttt{entity\_match} exist, where the LLM is making use of semantic similarity based entity matching tool to jump over the knowledge graph. After the Stage 3 RL training, more standardized tool call trajectories have emerged, where the LLM usually use \texttt{entity\_match} tool first to locate the starting node, then trigger a \texttt{node\_info} to get its relations with its neighbors. The standardized tool call trajectories reflects an adaptation of the LLM to the actual multi-hop question answering scenario, where the start node is explicit but the inner relations are supposed to be found from the edges. The standardization of the tool call trajectory could attribute to the reason of accuracy increase.

\subsection{Experiment summarization}

We summarize the key results from various stages of our model development, including baseline models and those trained with  supervised fine-tuning or reinforcement learning.

Unless otherwise specified, all base models referred to are Qwen3-14B. By default, these models operate in think mode, where the output consists of a reasoning segment (enclosed in \texttt{<think>} tags) followed by either a tool call or a final answer.

For the direct answering setting and the LightRAG (Qwen3-14B) baseline (reported in Table~\ref{tab:accuracy_wo_format_requirement}), no explicit answer format is enforced. Therefore, we adopt LLM-based evaluation using DeepSeek-R1 to assess answer correctness based on reasoning and content.

In contrast, for models that interact with knowledge graph tools (reported in Table~\ref{tab:accuracy_w_format_requirement}), the answer format is explicitly specified in the prompt. Consequently, we use exact match (EM) as the evaluation metric to assess the correctness of the final answer.

Note that EM-based and LLM-based evaluations are not directly comparable, as EM is stricter and penalizes deviations in output format even when the semantic content is correct.

\begin{table}[htbp]
\centering
\begin{tabular}{@{}cccc@{}}
\toprule
\textbf{Model/Agent} & \textbf{1/2 Hop (\%)} & \textbf{3+ Hop (\%)} & \textbf{Evaluation method} \\
\midrule
Direct answer & 26.0 & 26.3 & R1 \\
LightRAG & 62.5 & 49.5 & R1 \\
\bottomrule
\end{tabular}
\vspace{0.5em}
\caption{Qwen3-14B direct answer and LightRAG(Qwen3-14B) answer accuracy without format requirement}
\label{tab:accuracy_wo_format_requirement}
\end{table}

\begin{table}[htbp]
\centering
\begin{tabular}{@{}cccc@{}}
\toprule
\textbf{Model/Agent} & \textbf{1/2 Hop (\%)} & \textbf{3+ Hop (\%)} & \textbf{Evaluation method} \\
\midrule
Base model(before training) & 34.9 & 15.6 & EM \\
Rule-based RL(Stage 1, mix data) & 57.5 & 32.1 & EM \\
SFT(Stage 2, Lora,loss weight mask=0.001) & 89.2 & 60.5 & EM\\
SFT(Stage 2, Full,no think) & \textbf{94.1} & 68.4 & EM\\
Post-SFT RL(Stage 3) & 94.0 & \textbf{69.3} & EM\\
\bottomrule
\end{tabular}
\vspace{0.5em} 
\caption{Qwen3-14B reasoning with KG tools accuracies before/after training, with exact match being the evaluation method.}
\label{tab:accuracy_w_format_requirement}
\end{table}


\section{Related Works}

\label{sec_rel}

\textbf{Retrieval-Augmented Generation (RAG)} Classic “single-shot” RAG grabs one batch of passages (e.g., DPR + BART \cite{lewis2021retrievalaugmentedgenerationknowledgeintensivenlp}) and never looks back, which limits multi-hop reasoning. Newer methods turn retrieval into a loop. ReAct \cite{yao2023reactsynergizingreasoningacting} go further by letting the LLM decide when and what to trigger action, iterating until a stop signal, during which the tool call and result are preserved in the prompt context. Mindsearch\cite{chen2024mindsearchmimickinghumanminds} decomposes the complex question both in multi round and multi sub-query manner, leveraging a planning graph to trigger the web searcher for collecting extensive information then provides the final answer.
These multi-round strategies cut hallucinations and boost complex question answering accuracy, marking the shift from fixed one-shot retrieval to flexible, LM-driven evidence gathering.

\textbf{RL training on LLM reasoning with search tool} Several works have experimented on applying reinforcement leaning to LLM reasoning with search tool usage\cite{jin2025searchr1trainingllmsreason, zheng2025deepresearcherscalingdeepresearch,qian2025toolrlrewardtoollearning}. Most of them follows an interleaved pattern where the tool call triggered by LLM is followed by search results until the final answer is obtained. Search-R1\cite{jin2025searchr1trainingllmsreason} trained the LLM on 1/2 hop questions(NQ\cite{kwiatkowski-etal-2019-natural}, TriviaQA\cite{joshi-etal-2017-triviaqa}, HotpotQA\cite{yang2018hotpotqadatasetdiverseexplainable}, 2WikiMultiHopQA\cite{ho2020constructingmultihopqadataset} and etc.) with E5\cite{wang2024textembeddingsweaklysupervisedcontrastive} as retriever on a local corpus which is 2018 Wikipedia dump\cite{karpukhin2020densepassageretrievalopendomain}. DeepSearcher\cite{zheng2025deepresearcherscalingdeepresearch} trained on the similar 1/2 hop question datasets with question filtered based on some criteria, with open web search as the retriever.

\textbf{Natural Language to KG query translation}  
Some of the works approach the problem of knowledge graph querying by natural language to KG query translation. Text2cypher\cite{ozsoy2024text2cypherbridgingnaturallanguage} constructs a dataset consisting of question and corresponding cypher query, performs fine tuning on an LLM with the dataset and observes the performance increase in cypher query generation due to fine tuning. Other Works like \cite{zhao2023s2ctransbuildingbridgesparql}\cite{10.1145/3708326}\cite{DBLP:conf/i-semantics/KovriguinaTRM23} emphasize on Sparql query generation using LLM.

\textbf{LLM reasoning on knowledge graph} LLM reasoning on a knowledge graph can enable the LLM to trace on the knowledge graph and collect more information compared with one-go knowledge graph querying.
RoG \cite{luo2024reasoninggraphsfaithfulinterpretable} adopts a plan → retrieve → reason loop: the LLM sketches a relation-path plan, grounds it to the KG, and then reasons only on the retrieved path, yielding traceable, low-hallucination answers. Think-on-Graph \cite{sun2024thinkongraphdeepresponsiblereasoning} treats the LLM as an autonomous beam-search agent that iteratively expands and prunes entity/relation branches, using the discovered triplets as an explicit scratch-pad—no extra training required. KG-Agent \cite{jiang2024kgagentefficientautonomousagent} frames multi-hop KG-QA as a tool-augmented decision process: an instruction-tuned LLM selects actions (entity-link, relation-fetch, set-op, stop) until a termination criterion is met, achieving competitive accuracy with an order-of-magnitude less compute. 

\section{Conclusion}
\label{sec_conc}

This paper introduces a novel multi-stage training framework that mimics human behavior in handling multi-hop question answering on schema-free knowledge graphs. Our method shows substantial performance gains over baseline models, especially on complex questions involving 3+ hops. While we've validated our approach across various knowledge graphs, several areas remain to be explored. These include exploring scaling curves when applying the framework to larger models and integrating a progress reward model to ensure correct reasoning traces and prevent over-thinking.

\section*{Acknowledgements}
This project was completed from late April to early July in 2025. Special thanks to the Tanka AI team for the generous support in both time and computational resources.

\bibliography{main}

\appendix

\newcommand{\highlighttags}[1]{%
  \def\temp{#1}%
  \ifdefstring{\temp}{<think>}{\textcolor{red}{#1}}{%
  \ifdefstring{\temp}{</think>}{\textcolor{red}{#1}}{%
  \ifdefstring{\temp}{<tool\_call>}{\textcolor{blue}{#1}}{%
  \ifdefstring{\temp}{</tool\_call>}{\textcolor{blue}{#1}}{%
  \ifdefstring{\temp}{<answer>}{\textcolor{orange}{#1}}{%
  \ifdefstring{\temp}{</answer>}{\textcolor{orange}{#1}}{#1}}}}}}%
}

\section*{Appendix}
\addcontentsline{toc}{section}{Appendix}
\section{SFT Data Preperation}
\label{sec:sft-data-preparation}
Each data sample in the dataset is a JSON object with the following key fields:

\begin{itemize}
    \item \texttt{id}: A unique identifier for the sample.
    \item \texttt{graph\_type}: The domain of the knowledge graph (e.g., \texttt{agriculture}, \texttt{cs}). A full list of graph types is provided in Table \ref{tab:category_hop_distribution}.
    \item \texttt{question}: A natural language query that requires multi-hop reasoning over the knowledge graph.
    \item \texttt{golden\_answer}: The golden answer to the question.
    \item \texttt{conversations}: A list of multi-turn interactions capturing the full reasoning trace between the LLM and the environment, including tool calls and their observations. More details are provided below.
    \item \texttt{tools}: The tool schema embedded in the system prompt, describing the available tools and their parameters. The decription of tools are provided in Section \ref{sec:tools}.
    \item \texttt{system}: The instruction prompt provided to the model to guide its reasoning behavior and tool usage protocol.
\end{itemize}

The following is a sample of \texttt{conversations} field from our Supervised Fine-Tuning (SFT) dataset. It was constructed from multi-step reasoning outputs generated during reinforcement learning. Each step includes the CoT of the model, tool call, observations, and final answers. 

To align with supervised training requirements, we converted these multi-step outputs into a ShareGPT-style conversation format. Specifically, ShareGPT formatting requires that user messages such as prompts and tool observations appear at odd-numbered positions (e.g., ``from'': ``human'', ``from'': ``observation''), while assistant responses (model-generated content and tool calls) appear at even-numbered positions (e.g., ``from'': ``gpt''). 

To satisfy this alignment and preserve the logical progression of reasoning, we merged each internal reasoning cot with its corresponding tool call into a single assistant message. This unified message is attributed to ``from'': ``gpt'', representing the model's full thought-action step in one coherent output.

By following this structure, we are able to construct a clean, multi-turn training dataset where each round of reasoning and action is explicitly encoded. This format facilitates better instruction-following behavior and enables the model to learn tool-augmented reasoning in a controlled, consistent manner.

\vspace{1em}




\noindent\textbf{Tool Calling Trace:}

\begin{tcolorbox}[
    colback=white,
    colframe=black!40,
    title=Multi-turn Tool Calling Sample,
    fonttitle=\bfseries,
    breakable,
    width=1.1\textwidth,
    label={box:trace example},
]

\textbf{from: human}

\texttt{Which traditional grazing method in regional agriculture is practiced in the same region where Rational Grazing is compared in terms of grazing methods and regional practices?}

\hdashrule[0.5ex]{\linewidth}{1pt}{3pt}

\textbf{from: gpt}

\texttt{\highlighttags{<think>}}

\par\vspace{0.5em}\par

\texttt{To determine the traditional grazing method practiced in the same region as Rational Grazing, I need to identify the region where Rational Grazing is implemented and then find the traditional grazing method used there.}

\par\vspace{0.5em}\par

\texttt{\highlighttags{</think>}}

\par\vspace{0.5em}\par

\texttt{\highlighttags{<tool\_call>}}

\begin{verbatim}
entity_matcher(node_name="Rational Grazing", graph_type="agriculture")
\end{verbatim}

\texttt{\highlighttags{</tool\_call>}}

\hdashrule[0.5ex]{\linewidth}{1pt}{3pt}

\textbf{from: observation}

\texttt{Found 5 entities matching 'Rational Grazing':}
\par\vspace{0.5em}\par
\texttt{1. Entity: Rational Grazing Description: Rational Grazing Rational Grazing is a systematic and strategic approach to pasture and livestock management developed by André Voisin...}
\par\vspace{0.5em}\par
\texttt{2. Entity: Rational Grazing Principles Description: Rational Grazing Principles Principles outlined to guide systematic and effective grazing management...}
\par\vspace{0.5em}\par
\texttt{3...}
\par\vspace{0.5em}\par
\texttt{4...}
\par\vspace{0.5em}\par
\texttt{5...}

\hdashrule[0.5ex]{\linewidth}{1pt}{3pt}

\textbf{from: gpt}

\texttt{\highlighttags{<think>}}

\par\vspace{0.5em}\par

\texttt{Based on the information provided, Rational Grazing is a method that optimizes pasture and livestock management. To find the traditional grazing method practiced in the same region, I need to identify the region where Rational Grazing is applied and then look for traditional methods used in that area.}

\par\vspace{0.5em}\par

\texttt{\highlighttags{<think>}}

\par\vspace{0.5em}\par

\texttt{\highlighttags{<tool\_call>}}

\begin{verbatim}
node_info(node_name="Rational Grazing", graph_type="agriculture")
\end{verbatim}

\texttt{\highlighttags{</tool\_call>}}

\hdashrule[0.5ex]{\linewidth}{1pt}{3pt}

\textbf{from: observation}

\texttt{Entity: Rational Grazing... Related region: Pays de Caux}

\hdashrule[0.5ex]{\linewidth}{1pt}{3pt}

\textbf{from: gpt}

\texttt{\highlighttags{<think>}}

\par\vspace{0.5em}\par

\texttt{From the node information, it is evident that Rational Grazing has been applied in various regions, including North-West Europe, Africa, and Britain. Specifically, the entity "Pays de Caux" is mentioned in relation to Rational Grazing. Pays de Caux is located in North-West France.}

\par\vspace{0.5em}\par

\texttt{\highlighttags{</think>}}

\par\vspace{0.5em}\par

\texttt{\highlighttags{<tool\_call>}}
\begin{verbatim}
node_info(node_name="Pays de Caux", graph_type="agriculture")
\end{verbatim}
\texttt{\highlighttags{</tool\_call>}}

\hdashrule[0.5ex]{\linewidth}{1pt}{3pt}

\textbf{from: observation}

\texttt{Entity: Pays de Caux...}

\hdashrule[0.5ex]{\linewidth}{1pt}{3pt}

\textbf{from: gpt}

\texttt{\highlighttags{<think>}}

\par\vspace{0.5em}\par

\texttt{Based on the node information for "Pays de Caux", it is clear that Continuous Grazing is one of the traditional grazing methods practiced in this region. Since the question asks for the traditional grazing method practiced in the same region where Rational Grazing is compared, Continuous Grazing fits this description.}

\par\vspace{0.5em}\par

\texttt{\highlighttags{</think>}}

\par\vspace{0.5em}\par
\texttt{\highlighttags{<answer>}Continuous Grazing\highlighttags{</answer>}}

\texttt{}

\end{tcolorbox}

\definecolor{systembg}{RGB}{245,245,250}
\definecolor{systemborder}{RGB}{90,120,190}

\newpage
\section{System Prompt Format and Tool Description Example}
\label{sec:systemp_prompt}

\begin{tcolorbox}[colback=gray!5!white, colframe=gray!80!black, title=System Prompt:, fonttitle=\bfseries]
\texttt{Answer the given question using knowledge graph tools.
You should use an iterative process of reasoning and tool usage to solve the problem. For each step:}
\vspace{1em}  

\texttt{1. First, think about what you know and what information you need by writing your thoughts inside \textless think\textgreater{} and \textless /think\textgreater{} tags.}

\texttt{2. You can only use one tool at a time}

\texttt{3. After receiving tool results, think again about what you've learned and what to do next}

\texttt{4. Repeat this process of thinking and searching until you have enough information to answer the question.}

\texttt{5. When you have enough information to answer the question, represent your final answer between \textless answer\textgreater{} and \textless/answer\textgreater{} tag.}

\vspace{2em}

\texttt{\# Tools}

\vspace{1em}
\texttt{You can call only one function at one time to assist with the user query. You are provided with function signatures within \textless tools\textgreater{} and \textless/tools\textgreater{} XML tags:}

\vspace{1em}

\texttt{\textless tools\textgreater{}}

\texttt{\{"type": "function", "function": \{"name": "entity\_matcher", "description": "A tool that Finds entities in the knowledge graph that match or are similar to your query", "parameters": \{"type": "object", "properties": \{"node\_name": \{"type": "string", "description": "The entity or concept you want to search for (e.g., `crop diseases`, `legal precedent`)"\}, "graph\_type": \{"type": "string", "description": "Type of knowledge graph to query, must be `\{\{ graph\_type \}\}` for this question", "enum": ["\{\{ graph\_type \}\}"]\}\}, "required": ["node\_name", "graph\_type"]\}\}\}} \\

\vspace{1em}

\texttt{\{"type": "function", "function": \{"name": "node\_info", "description": "A tool that retrieves detailed information about a specific entity and its relationships", "parameters": \{"type": "object", "properties": \{"node\_name": \{"type": "string", "description": "The exact entity name (use names returned by entity\_matcher)"\}, "graph\_type": \{"type": "string", "description": "Type of knowledge graph to query, must be `\{\{ graph\_type \}\}` for this question", "enum": ["\{\{ graph\_type \}\}"]\}\}, "required": ["node\_name", "graph\_type"]\}\}\}\}} \\
\texttt{\textless/tools\textgreater{}} \\

\vspace{1em}

\texttt{For each function call, return the call in Python function-call style within <tool\_call></tool\_call> tags, for example:} \\
\texttt{<tool\_call>node\_info(node\_name="Rastrigin Function", graph\_type="cs")}
\texttt{</tool\_call>}

\end{tcolorbox}

\newpage
\section{Multi-hop question generation and evaluation prompt}

\label{sec:question_generation_prompts}

\begin{tcolorbox}[colback=gray!5!white, colframe=gray!80!black, title=1/2 hop question generation prompt, fonttitle=\bfseries]

\texttt{Relations(entity\_1 - entity\_2: the relation):}

\texttt{\textasciigrave\textasciigrave\textasciigrave}

\texttt{Urban Farms - Corboy: perception change, urban development}

\texttt{...}

\texttt{\textasciigrave\textasciigrave\textasciigrave}

\texttt{The above is a knowledge graph where the relation between the entities are listed.}

\texttt{Your need to determine a path in the knowledge graph first, and the path should be suitable for generating a one-hop or two-hop question.}

\texttt{The suitable path should be:}

\texttt{The language of the one/two hop question will be natural.}

\texttt{The answer will be unique, only one node in the knowledge graph corresponds to the correct answer.}

\texttt{The length of the path is 2 or 3, which corresponds to one-hop and two-hop questions.}

\texttt{Then you need to generate a one/two hop question from the path. You should not reveal the answer node name but need to provide the starting node name(s) for the answerer to start with. Use natural language to describe the relation between the nodes.}

\texttt{Here are some two-hop question types as hint:}

\texttt{1. Inferring the bridge entity to complete the 2nd-hop question}

\texttt{Which team does the player named 2015 Diamond Head Classic’s MVP play for?}

\texttt{2. Locating the answer entity by checking multiple properties}

\texttt{Which former member of the Pittsburgh Pirates was nicknamed ”The Cobra”?}

\texttt{3. Inferring about the property of an entity in question through a bridge entity}

\texttt{What city is the Marine Air Control Group 28 located in?}

{\small\color{gray}
\texttt{These are the path and questions generated before but they are not good enough.}

\texttt{previous path: ['Corboy', 'Beehives', 'Greensgrow']}

\texttt{previous question: Through his personal commitment to beekeeping with Beehives...?}

\texttt{This is the feedback}

\texttt{path validity: True}

\texttt{question feedback: The question reveals the name of a middle node 'Beehives' in the path, which should be hidden according to the requirements.}

\texttt{You need to provide a better question and/or path and fix the previous issues.}
}

\texttt{Output in English.}

\texttt{Output in JSON format only, don't add markdown code delimiter or anything else, only a valid json object:}

\texttt{\{"path": [node\_name1, node\_name2, ...], "question": "question", "answer": the answer node name\}}

\end{tcolorbox}

\newpage
\begin{tcolorbox}[colback=gray!5!white, colframe=gray!80!black, title=3+ hop question generation prompt:, fonttitle=\bfseries]

\texttt{Relations(entity\_1 - entity\_2: the relation):}

\vspace{1em}

\texttt{\textasciigrave\textasciigrave\textasciigrave}

\texttt{Farmyard Manure - Humus: soil nutrition, organic matter transformation}

\texttt{Soil Respiration - Humus: soil process, nutrient cycling}

\texttt{...}

\texttt{\textasciigrave\textasciigrave\textasciigrave}

\vspace{1em}

\texttt{Given the knowledge graph above, identify one path that is suitable for generating a multi-hop question. A suitable path should meet the following criteria:}

\texttt{The resulting question should be natural and fluent in language (not awkward or difficult to read).}

\texttt{The answer should be unique and unambiguous.}

\texttt{The path length should be at least 4.}

\vspace{1em}

\texttt{Once such a path is selected, generate a multi-hop question based on it. The question must follow these constraints:}

\texttt{Only the name of the start node or end node in the path may be explicitly mentioned. The names of the intermediate nodes must not be revealed, as doing so would reduce the difficulty of the question.}

\texttt{The relationships between each pair of connected nodes should be described using natural language.}

\texttt{Formulate one comprehensive multi-hop question, rather than multiple smaller sub-questions.}

\vspace{1em}

{\small\color{gray}
\texttt{These are the path and questions generated before but they are not good enough.}

\texttt{previous path: ['Sir Albert Howard', 'Humus', 'Soil Fertility', 'Forest', 'Soil Erosion']}

\texttt{previous question: Research associated with Sir Albert Howard emphasizes ...?}

\texttt{This is the feedback}

\texttt{path validity: False}

\texttt{question feedback: The question explicitly mentions 'Sir Albert Howard' as the starting node, ... }

\texttt{You need to provide a better question and/or path and fix the previous issues.}
}

\vspace{1em}

\texttt{Output in English.}

\vspace{1em}

\texttt{Output in JSON format only, do not add markdown code delimiter or anything else, only a valid json object:}

\texttt{\{"path": [node\_name1, node\_name2, node\_name3, node\_name4, ...], "question": "question", "answer": the answer node name\}}

\end{tcolorbox}

\newpage

\begin{tcolorbox}[colback=gray!5!white, colframe=gray!80!black, title=Multihop question evaluation prompt:, fonttitle=\bfseries]

\texttt{You will see a path from knowledge graph and a corresponding multi-hop question. Your task is to check if the question satisfies all the following requirements:}

\vspace{1em}

\texttt{The question contains the name of the node explicitly so that the answerer knows where to search from in the knowledge graph.}

\texttt{The question only reveals the name of node to start from, but hide the name of the other nodes in the path.}

\texttt{The question is a multi-hop question.}

\vspace{1em}

\texttt{If the question satisfies all the requirements, the question is valid, otherwise it is invalid.}

\vspace{1em}

\texttt{Here is the real data:}

\texttt{Path: ['Miller', 'North Dakota', 'California', 'Central Valley', 'Almond Pollination']}

\texttt{Multi-hop question: What agricultural support process follows seasonal beekeeping migration to a northern state, involves competitive honey production in a western state, and occurs in a valley region providing crop pollination services?}

\vspace{1em}

\texttt{Output in JSON format only, don't add markdown code delimiter or anything else, only a valid json object:}

\texttt{\{"valid": true or false, "reason": "your reason"\}}

\end{tcolorbox}

\end{document}